\documentclass{article}
\usepackage[utf8]{inputenc}

\usepackage{graphicx}
\usepackage{authblk}
\usepackage{color}
\usepackage{url}
\usepackage{amsmath}
\usepackage{MnSymbol}
\newtheorem{definition}{Definition}
\newcommand{\R}{{I\!\! R}}
\usepackage{appendix}

\title{The Complexity of Comparative Text Analysis -- ``The Gardener is always the Murderer'' says the Fourth Machine}
\author[1]{M. Weber}
\author[2,1]{K. Fackeldey}
\affil[1]{Zuse Institute Berlin (ZIB), Takustra{\ss}e 7, 14195, Berlin, weber@zib.de, fackeldey@zib.de}
\affil[2]{Technische Universit{\"a}t Berlin (TUB), Mathematik, Stra{\ss}e des 17. Juni 136, 10623 Berlin}
\date{\today}

\begin{document}

\maketitle

\begin{abstract}
    There is a heated debate about how far computers can map the complexity of text analysis compared to the abilities of the whole team of human researchers. A ``deep'' analysis of a given text is still beyond the possibilities of modern computers. 
    In the heart of the existing computational text analysis algorithms there are operations with real numbers, such as additions and multiplications according to the rules of algebraic fields. However, the process of "comparing" has a very precise mathematical structure, which is different from the structure of an algebraic field. The mathematical structure of "comparing" can be expressed by using Boolean rings. We build on this structure and define the corresponding algebraic equations lifting algorithms of comparative text analysis onto the ``correct'' algebraic basis. From this point of view, we can investigate the question of {\em computational} complexity of comparative text analysis. 
\end{abstract}

\section{Introduction}

Some texts are fascinating. Their authors seem to capture the zeitgeist in a very precise and emotional way. Some texts are innovative, densely written, and motivating us to deeper think about unexpected abstract relations. Other texts, however, seem to be outdated. Even their grammar and style seem to be out of our comfort zone. All presented scenes are only mixtures of old ideas. The authors do not ``touch'' or ``catch'' us. This could be seen as a non-scientific way of comparative text analysis. 

Scientific comparative text analysis is beyond ``like and dislike''. It can work out the scribal skills of authors, their connection and position within their own epoch. It also works out the change of language and grammar, or the acquisition of vocabulary of civilizations that trade with one another. 

Linguistic and literary approaches need thus to be differentiated. The field of modern linguistics, rooted in Saussure’s definition of langue as a system and parole as its actualization 
in specific speech acts \cite{Sau1959}, deals with linguistic categories such as phonetics, morphology, lexis, syntax, etc. Following Saussure, the study of this system is bound by paradigms and co-occurrence patterns of the entities involved, which can also be quantified. The study of these quantifiable variations in texts is central to the field of corpus linguistics (cf. \cite{Biber1998, Biber2009}), which draws upon a representative and principled set of texts. Quantitative Methods such as multivariate statistical analyses are widely used in corpus linguistics, see for example an application of cluster analysis in respect to the Tyneside dialect of English \cite{Moisl2015}.

The situation is different in literary studies. In her article \cite{DHB} “The digital humanities debacle“, the literary scholar Nan Z. Da describes the problem of constructing a machine that can perform text analysis. Machines mostly ignore the complexity of qualitative text analysis and are mostly restricted to apply statistical results about the ``content'' of the texts. Evelyn Gius \cite{Gius} answered to this criticism and describes five dimensions of complexity of text analysis. From our point of view, she addresses some problems of computation which are also known from numerical mathematics. In this regard, there are two different meanings of the term ``comparative text analysis'':

\begin{itemize}
    \item[TA] In order to analyze one given text, it is often mandatory to take into account different other texts (from the same epoch, the same author, from history books, or just texts from the same register). The important aspects within a text analysis can only be defined by comparing the given text with other texts. In the case of TA, we are only interested in an analysis of the one single text. This would be denoted as {\em text analysis} in this manuscript. Writing an ``essay'' about one single text always takes into account our own experiences and our own knowledge base. The acquisition of different text sources in order to analyse {\em one} given text is included in the mapping $f$, which will be defined in Section~\ref{sec:ring}. It is defined what TA is. It is not defined how TA has to be done. 
    \item[CTA] The other meaning of comparative text analysis is to figure out differences and commonalities between given texts. Usually, after we have had (or better: while we have) a deeper TA-look at the single texts. Here the research question is to find out what characterizes or separates one group of texts from another group of texts. In mathematics this problem of finding the ``groups'' and their characteristics is denoted as {\em clustering} \cite{clustering}. The manuscript is about this kind of comparative text analysis.
\end{itemize}

The term ``complexity'' also has a lot of different meanings. As described in TA, the data which is given by the texts alone is not enough. Without taking further sources into account, the analysis of texts is not complete. This is the problem of {\em non-existence of a solution}. The outcome of the analysis is not determined by the input text alone. Gius also addresses the problem of {\em instability} (thinking of TA as an algorithm). Some questions concerning texts (e.g., ``Is the main character in a novel ill?'') are difficult to answer. The author can write hints about the characters in such a way, that the reader is undecided about some aspects of interests. If a machine tries to give a $\{0,1\}$-answer, the decision depends on very subtle changes to the text or is even {\em non-unique}. In principle (with each of these three observations), Gius implies that TA is an {\em ill-posed} problem according to Jacques Hadamard \cite{hadamard}. Especially since the 1950s, mathematicians have been concerned intensively with how to make the solution of this kind of problems "computable". Particularly noteworthy are the works of Russian mathematicians, which increasingly developed into a coherent theory which has been presented in the 1990s \cite{tikhonov}. The mathematical concept of ``regularization'' is one standard approach to deal with ill-posed problems. In principle, regularization restricts the set of possible answers about a given problem, which will be a guiding idea of this manuscript, too.\\
Let us consider an example for the ill-posedness of text analysis in the sense of Hadamard. This is the text:
\begin{center}
\textit{Gauß once said: "There are three kinds of mathematicians, the one who can count and the ones who can't."}    
\end{center}
In this text it is difficult to answer questions like: ``Is there a typo in the text?'', since is involves knowledge beyond this text. A search for ``irregularities'' in this text would uncover that either the word ``three'' is wrong or the enumeration is incomplete, which is by the way, a correct analysis of the text. However a further analysis in a wider context would uncover, that Gauß himself was a mathematician. Either he did indeed a mistake or the whole text is meant as a joke or a witty justification of his own calculation error, which also expresses that a mathematician of his reputation deals with deeper thoughts than with counting (i.e., ``three'' is not a typo). In this context the problem of text analysis is not well posed because it has different levels of answers which are correct on their corresponding level on analysis, including information that is not given by the text itself. However, we quickly recognize the joke behind this text, as we can recognize this type of construction of contradictions ({\em compared} to other jokes, if it meets our sense of humor).

Regarding this ill-posedness, we first postpone the problems of computer-based text analysis in a very easy way: We allow literary scholars to write down whatever they want. We even allow for contrary opinions about texts, because we do not ``define'', how texts should be analyzed. Like in numerical mathematics we start with a different question: What kind of problem has to be solved? Among all aspects of comparative text analysis that can hardly be concretely "quantified" there is, however, one unchangeable and concretely definable action - namely the action of "comparing". CTA is a clustering problem. There exist mathematical algorithms to carry out  a clustering, whenever the input data is ``readable'' by  a mathematical formalism.  Thus, first of all we introduce a mathematical formalism for solving clustering problems in Section \ref{sec:ring}. Later on, we will also formulate some algorithms to solve clustering problems in Sections \ref{sec:PCCA+} and \ref{sec:kernel}. The implicit answer to ill-posedness of text analysis is here given by a restriction of the set of possible solutions. 

Having defined the mathematical problem,  what is the complexity of this problem? How many operations are needed to solve the problem depending on the size of the input data? This kind of analysis will be given at the end of this manuscript in Section \ref{sec:NP}. It is the question of {\em computational} complexity. This kind of complexity analysis is currently much discussed in the mathematical community and leads to the formulation of a research question (${\cal NP}=\cal P$?), which still seems unsolved today and for the answer to which one can actually win a million dollars. More about this in Section \ref{sec:NP}.  

``Complexity'' can also be mathematically understood in a different way. Our computations are mappings. Mappings from input data to some output data. ``Complexity'' is also a property of this mapping. Assume, we have constructed an algorithm to solve our problem. Now, we apply this algorithm to some input data (not to all possible input data). If our choice of input data always leads to the same output, then this mapping is not complex (although, a silly algorithm can perform a lot of computational steps to always give the same answer). The complexity of the CTA mapping will be investigated in Section \ref{sec:complex1}.  

Although we have postponed the complexity problems of TA by focussing on the clustering, comparative text analysis is not independent from text analysis.  In Section \ref{sec:central} we ignore our good intentions (regarding the digital humanities debacle) and define how to do text analysis on the computer. The central idea in this section is: One can eventually learn to do text analysis if one understands what is important or interesting when comparing texts. On the one hand, this forms the basis for actually being able to implement clustering algorithms one day, because it explains what kind of algebraic transformations have to be implemented in the machines. On the other hand, it subordinates existing quantitative text analysis to our concept, see also Sec.~\ref{sec:basis}.

Gius mentions in her article \cite{Gius} a further problem of, e.g., training an artificial intelligence to do TA. It is a {\em statistical problem}. Compared to the high complexity of texts, there are only a few text sources  available for learning text analysis. A possibility to circumvent this problem is implicitly presented in this manuscript. Here, we break down the process of comparative text analysis into a few individual operations. We construct four separate machines for these individual operations. These machines can be combined in different ways.  The capabilities of these special machines can be accessed by whether they achieve high quality results in different compositions. This quality can be checked with the available text inputs. From this point of view, the abundance of training objects is given by the abundance of compositions of the machines and not only by the abundance of texts. Two different compositions are mentioned in Section \ref{sec:implicit}. In a less mathematical formulation: In order to train comparative text analysis, not the high number of available texts is important, but being able to answer many different research questions with regard to the given data base. This will be exemplified in Fig.~\ref{arch}. 

The final dimensions of Gius' complexity model relate to whether computer-based text analysis really makes a contribution to the gain in knowledge. She says, that this can be seen in whether the machines can produce results without human help and whether these results have an impact.

In terms of being able to put the (human) trained machines into different compositions, the question of knowledge gain is connected to the question of functionality of the machines. Do they produce reasonable, interesting results (without human help) after rearrangement into new compositions? Do they produce meaningful clusterings?

\section{The rough course of comparative text analysis}\label{sec:basis}

\begin{figure}[ht]
\centering
\includegraphics[width=0.8\textwidth]{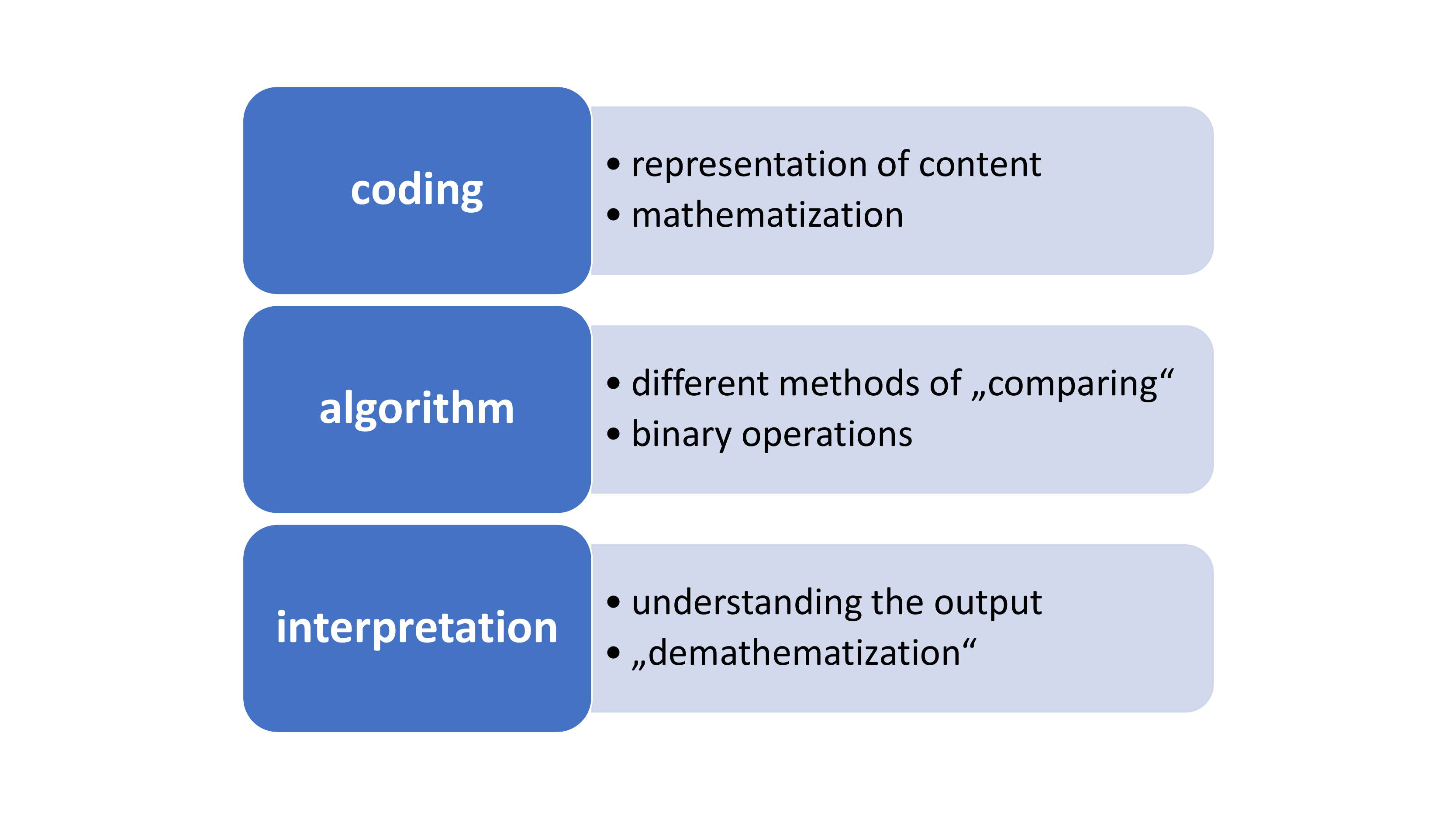}
\caption{\label{fig} Three steps of computational comparative text analysis.}
\end{figure}

This section is about the current computational approach to comparative text analysis. In principle, one can subdivide this approach into three consecutive steps, see Fig.~\ref{fig}. 

\paragraph{Coding.} The first step turns the texts into something that can be treated and processed by a computer (or at least by mathematics). Instead of treating the texts themselves, they and their contents are just turned into mathematical objects (e.g., real-valued vectors). The coding will be expressed by a mapping $f:S\rightarrow {\cal B}$, where $S$ is the set of all possible texts and $\cal B$ is the set of representations of these texts. In Sec.~\ref{sec:ring} it will be explained what these representations are. In principle, $f(a)$ represents the {\em whole} content, structure, interpretation margin, ... of the text $a$. 

\paragraph{Algorithm.} After the texts are coded and, thus, available for computations (e.g., clustering algorithms), computational methods are applied. Computations are based on binary operations. In essence, these operations can be traced back to additions and multiplications. The algorithms in this manuscript are also based on binary operations -- binary operations on the elements of $\cal B$, which will be defined in Sec.~\ref{sec:algebra}. This section will present one central idea of this manuscript: The algebra of "comparing" is different from the algebra of fields. 

\paragraph{Interpretation.} The outcome of the algorithm is a mathematical object. In order to provide a meaning with this object, we need to interpret the outcome. The interpretation of the algorithmic results is again a mapping. The output of the algorithmic part (elements of $\cal B$) has to be transformed. In this manuscript, this transformation will be defined in Sec.~\ref{sec:implicit} as $\rho:{\cal B}\rightarrow S$. The interpretation is formulated as a text. 

\section{Construction of a Boolean Ring}\label{sec:ring}

Imagine $S$ is the set of all possible texts. Texts in any (old , current, future) language (even with mixed languages) with any way of using letters or coding. Texts that have already been written, texts that will be written or {\em can} be written. Texts which are arbitrarily long. Parts of texts are texts and even (yet) senseless combinations of letters are texts. Who has guessed that "{\#}JKvsP7" will ever have a certain meaning? 

\subsection{Text analysis is a mapping}
$S$ is assumed to be an infinite {\em set} of texts. Although, later on, we will restrict our studies to finite subsets of $S$. Now take one element $s$ of this set $s\in S$, e.g., the poem ``Prometheus''  by Johann Wolfgang von Goethe. The interesting thing about text analysis is: When writing a text analysis about ``Prometheus'' the outcome is again a text. There are many different possible analyzes that people can write about that poem. This {\em subset} of $S$, which includes all texts that can be written when analyzing $s$ is denoted as $f(s)\in {\cal B}$, where ${\cal B}={\cal P}(S)$ is the power set (the set of all subsets) of $S$. In this sense, text analysis is a mapping $f:S \rightarrow {\cal B}$. Every text $s$ is mapped to the set of possible ``essays'' about $s$, denoted as $f(s)$. Thus every text $s\in S$ maps to a subset $f(s)\in{\cal B}$. However, not every subset in ${\cal B}$ can be seen as possible ``secondary literature'' for an existing text. The elements of $\cal B$ are simply the subsets of $S$. The element $0\in{\cal B}$ denotes the empty set. 

When a person (e.g., a philologist) writes a comparative analysis of two texts $a$ and $b$, then in principle the following happens. First of all, the subsets of possible essays are created $A=f(a)$ and $B=f(b)$. If the person wants to write about the commonalities of the two texts, the written analysis will stem from the intersection of the two sets $A$ and $B$. This leads to a set $C=A\odot B$, where $\odot:{\cal B}\times {\cal B}\rightarrow {\cal B}$ denotes the intersection. If the person wants to write about the differences of $a$ and $b$, then an element of $D=A\oplus B$ is searched for, where $\oplus:{\cal B}\times {\cal B}\rightarrow {\cal B}$ denotes the symmetric difference of the two subsets $A$ and $B$, i.e., all elements which are in $A$ but not in $B$ and also all elements of $B$ which are not in $A$. 

\subsection{Algebraic structure of $\cal B$} \label{sec:algebra}
There are many interesting equations which are based on these operations $\odot$ and $\oplus$ when thinking of comparative text analysis. Instead of listing all these equations, mathematicians worked out the algebraic structure  of ${\cal B}$. They know that ``$({\cal B}, \oplus, \odot)$ is a Boolean ring''.  A Boolean ring is a special algebraic ring. Let us start with those interesting equations of comparative text analysis which imply the structure of an algebraic ring: 

\begin{definition} \cite{Dedekind, Hilbert, Noether}
Given a set $\cal B$ and two binary operations $\oplus:{\cal B}\times {\cal B}\rightarrow {\cal B}$ and $\odot: {\cal B}\times {\cal B}\rightarrow {\cal B}$. Then $({\cal B},\oplus, \odot)$ is denoted as an {\em algebraic ring}, if the following conditions hold for all (not necessarily pairwise different) elements $A,B,C\in {\cal B}$:
\begin{itemize}
    \item[(i)] the laws of distribution: $A\odot(B\oplus C)=(A\odot B) \oplus (A\odot C)$ and\\ $(B\oplus C)\odot A=(B\odot A) \oplus (C\odot A)$,
    \item[(ii)] the associative law: $(A\odot B)\odot C=A\odot(B \odot C)$, and
    \item[(iii)] that $({\cal B},\oplus)$ is a commutative group, i.e., 
    \begin{itemize}
        \item[a)] the associative law holds: $(A\oplus B) \oplus C=A\oplus (B \oplus C)$,
        \item[b)] commutivity holds: $A \oplus B=B\oplus A$, 
        \item[c)] there exists an element $0\in\cal B$, such that $0\oplus A=A$ for all $A\in \cal B$, and
        \item[d)] for every $A\in\cal B$ there is an element $\overline{A}\in \cal B$ such that $A\oplus \overline{A}=0$.
    \end{itemize}
\end{itemize}
\end{definition}

The definition of an algebraic ring does not include all algebraic expressions which we could write down for comparative text analysis. There is one important further equation. It is the idempotency $A \odot A= A$ which additionally holds and which turns the algebraic ring into a Boolean ring. 

\begin{definition} \cite{BooleR}
An algebraic ring $({\cal B}, \oplus, \odot)$ is denoted as {\em Boolean ring}, if idempotency $A \odot A= A$ holds for every $A\in {\cal B}$.
\end{definition}

The definition of an algebraic ring or of a Boolean ring does not include a neutral element of multiplication, i.e., we not necessarily have to assume an element $1\in {\cal B}$ with $1\odot A=A$ for all $A\in{\cal B}$. A ring which has such an element $1$ is called a {\em ring with unity}. In our case, the complete set of texts $S\in{\cal B}$ has this role, i.e., $1=S$. In this sense, the expression $B=1\oplus A$ means, that we create the subset $B$ of all texts which are {\em not} element of $A$. $B$ is the complement of $A$.

\subsection{Implications of $A\odot A=A$}

Boolean rings (and Boolean algebras) are studied in complexity analysis, computational algebra, and in computer science. The arithmetic laws formulated in the two definitions can be used to transform equations. From idempotency some further properties of Boolean rings can directly be derived. For instance, the equation $A\oplus A=0$ formalizes that there is nothing to be written when we want to figure out the differences between $A$ and $A$. In other words the element $\overline{A}$ in item (iii d) of the definition of an algebraic ring is equal to $A$ in comparative text analysis. The equation $A\oplus A=0$ does not occur in the definition of a Boolean ring, because it is already a consequence of $A\odot A=A$.  This can be shown in the following way: $(A\oplus A)\odot(A\oplus A)=A\oplus A$ by the idempotency. Furthermore, $(A\oplus A)\odot(A\oplus A)=A\oplus A\oplus A\oplus A$ by the law of distribution. Thus, $A\oplus A\oplus A\oplus A=A\oplus A$, which shows $A\oplus A=0$.  

Also the commutative law $A\odot B=B\odot A$ is a consequence of $A\odot A=A$ and of $A\oplus A=0$. Note, that $A\oplus B = (A \oplus B)^2=A\oplus (A\odot B) \oplus (B\odot A) \oplus B$. This means $(A\odot B) \oplus (B\odot A)=0$, which proves the commutative law of multiplication.  Boolean rings are commutative rings.  More precisely we have 
\[
\begin{split}
A\odot B & = (A\odot B)   \oplus \underbrace{((B\odot A) \oplus (B\odot A))}_{=0}\\
         & = \underbrace{((A\odot B)   \oplus (B\odot A))}_{=0} \oplus (B\odot A) \\
         & = B \odot A.
\end{split}
\]

\subsection{Efficiently build on existing literature}\label{sec:LR}
{Research questions from comparative text analysis are often of the form: ``Here we have two stacks of texts (of course we think of the possible essays about them). The left stack is  $X_1,\ldots , X_k\in{\cal B}$ and the right stack is $X_{k+1},\ldots, X_n\in{\cal B}$, where $1<k<n$. What are the characteristics of the left stack, that differentiate the texts from the right stack?'' }

This kind of questions occurs (implicitly or explicitly) in many applications of comparative text analysis. Just to give two simple examples, where this kind of clustering is applied: 

\begin{itemize}
    \item[Ex1.] In order to identify plagiarism, the sudden change of style of a  written text can indicate that this text is not stemming from only one author \cite{Alzahrani}. Finding this change of style is a clustering problem. What differentiates the first part of the text from the second part?  
    \item[Ex2.] In their Introduction \cite{power} G{\"o}hler et al. write: {\em With language we bring the world "to the point" with it we largely regulate social relations. Language is a condition for exercising power and is itself a power - power understood as "that factor in a social relationship that structures the actors' options for action“}. From this point of view, a change of language can indicate a change of rulers. What differentiates the use of language before some historical event compared to the language after this event?     
\end{itemize}

Let us reformulate the problem of finding differences in an algebraic way. The common properties of the left stack are given by $L=X_1 \odot \ldots \odot X_k$. Not the whole set $L\in{\cal B}$ is interesting for us. We have to exclude the essays that could be written about $X_{k+1}$ and those for $X_{k+2}$ and so forth.  Thus, we have to  multiply $L$ with $R=(1\oplus X_{k+1})\odot \ldots \odot (1\oplus X_n)$.

As an example: We have two stacks of poems. The left stack is ``Phenomenal Woman'' (by Maya Angelou) and ``If You Forget Me'' (by Pablo Neruda). On the right stack we have ``I carry your heart with me'' (by Edward Estlin Cummings) and ``The Road Not Taken'' (by Robert Frost). $X_1=f(\mathrm{``Phenomenal\, Woman"})$ is the set of all possible essays about Maya Angelou's poem. In a similar way, $X_2, X_3,$ and $X_4$ are defined. The task to figure out what are the characteristics of the left two poems which are not characteristics of the right two poems is the same task as to compute $L\odot R=X_1\odot X_2\odot(1\oplus X_3)\odot(1\oplus X_4)$. Algebra tells us, that we can solve this task in very different ways. It tells us {\em all} possible ways to solve it, simply by transforming the equation $L\odot R$. One possibility is $$L\odot R=(X_1\odot X_2)\oplus(X_1\odot X_2\odot X_3)\oplus (X_1\odot X_2\odot X_4)\oplus (X_1\odot X_2\odot X_3\odot X_4).$$    
We will have to find out certain commonalities of certain subsets of the four poems and then to work out the differences between our findings.  Alternatively, 
$$L\odot R=(X_1\odot X_2)\odot(1\oplus X_3\oplus X_4 \oplus(X_3\odot X_4)),$$
where the expression $X_3\oplus X_4 \oplus(X_3\odot X_4)$ is the union of $X_3$ and $X_4$.

Thus, research questions from comparative text analysis can be transformed into
algebraic equations. Using the arithmetic laws of a Boolean ring, we can transform these equations. But does it help? Do these transformations add any new insight into comparative text analysis?

So far, we presented an example for a possible task in comparative text analysis, but we did not provide {\em any} method or procedure to actually analyze the given poems. Mathematics is not taking over the task of text analysis from literary studies. However, mathematics can help to organize the work that has to be done in order to solve the given research question. Imagine, we find a lot of common secondary literature about ``I carry your heart with me''  and ``The Road Not Taken'', then solving the task to compute $L\odot R$ makes (at best) use of the term $X_3\oplus X_4 \oplus(X_3\odot X_4)$. In contrast to this: If we find a lot of literature about the commonalities of all four poems, then maybe the first expression to compute $L\odot R$ is more practical. 

Mathematics addresses the complexity of the problem and searches for efficient ways to solve it.

\section{Strategies of Automation}


During the history of mathematics, we invented the place value system, such that a machine with a finite storage capacity can in fact carry out certain calculations with real numbers. A multiplication like $f(\mathrm{``Phenomenal\, Woman"})\odot f(\mathrm{``If\, You\, Forget\, Me"})$, i.e., generating {\em all} possible common texts that can be written about the two poems, is impossible for a machine with a finite storage. However, there has been mathematics long before the computer has been invented, there have been quantum-algorithms \cite{shor} before the quantum computer has been built. We can study the algebra of text analysis, before we invent a machine to perform these tasks. In this section, we will try to find strategies to automate this algebra.

\subsection{Restrictions}
Not only from a computational perspective but also from the perspective of comparative text analysis, it is impossible to figure out {\em all} aspects that could be written down about two texts. {When writing a comparative text analysis we restrict ourselves to a certain list of aspects, e.g., analysis of the grammar, the style, the wording, the atmosphere, the reproduction of gender roles...} Thus, these restrictions cut out a certain (maybe finite and manageable) part of $S$. Not {\em every} text that could possibly be written will be written. Instead of taking into account the whole set $S$, we restrict ourselves to a subset $T\in{\cal B}$. This means, instead of applying the algebraic rules to $X_1=f(\mathrm{``Phenomenal\, Woman"})$, we apply them to $T\odot X_1$. In principle, we multiply every element of $\cal B$ with $T$. The element $X$ turns into $\phi(X)=T\odot X$, where the mapping $\phi:{\cal B}\rightarrow {\cal B}'$ restricts our set $\cal B$ to a certain subset $\cal B'$. This subset ${\cal B}'={\cal P}(T)$ restricts our studies to certain aspects of comparative text analysis. Furthermore, $\cal B$' is a Boolean ring with unity $T\in{\cal B}'$. $T$ will therefore be denoted as $1$ with regard to $\cal B'$. Restricting our considerations to a certain subset of texts $T$ does not destroy the properties of a ring. Thus, our arithmetic rules carry over to this subset. Additionally, the mapping $\phi$ has a very interesting algebraic structure, which is essential for turning our theoretical concept into a practical concept. 

\begin{definition} (cf. \cite{bosch})
Let $\cal B$ and $\cal B'$ be rings with unity. A mapping $\phi:{\cal B}\rightarrow {\cal B}'$ is denoted as {\em ring homomorphism} if the following holds for all $A,B \in{\cal B}$:
\begin{itemize}
\item[(i)] $\phi(A\oplus B)=\phi(A)\oplus\phi(B)$, 
\item[(ii)] $\phi(A\odot B)=\phi(A)\odot \phi(B)$, and
\item[(iii)] $\phi(1)=1$. 
\end{itemize}
\end{definition}

In the special case of our Boolean ring $\cal B$, the mapping $\phi(X)=T\odot X$ is a ring homomorphism: (i) is a consequence of the law of distribution, (ii) is a consequence of idempotency and the commutative law of $\odot$. (iii) is a little bit more difficult to see, because $\phi(1)=T$. However, $T$ is the unity in the ring $\cal B'$, because it is the ``full'' set of texts in this ring.  

The fact, that $\phi$ is a ring homomorphism has an important meaning. We already knew that the arithmetic rules of intersection and symmetric differences carry over to $\cal B'$. Being a ring homomorphism, the arthmetic rules carry over {\em element-wise} and the way to ``identify'' elements of $\cal B'$ with elements of $\cal B$ is given by $\phi$.

\subsection{Explicit restrictions}

{We have already described, how a problem of comparative text analysis is formulated in terms of an algebraic problem.


In order to fully automate the solution procedure, all calculations have to be performed by the machine. This will lead to a machine that would ``define'' how comparative text analysis has to be done. Thus (having the digital humanity debacle in mind), we should not trust the output of that machine.}

The interesting observation is, that mathematicians also do not trust the output of $+$ and $\cdot$ calculations of computers, because they know that real numbers are not ``perfectly'' represented in their machines. The machines produce errors. The whole research field of numerical analysis is based on how to handle these errors (cf. \cite{dd}).    

The proposed solution of comparative text analysis is to restrict the set $S$ to a manageable subset $T$. From an algebraic point of view, there are no further side constraints to be taken into account when choosing a restriction $T$. Any restriction leads to a ring homomorphism $\phi$. 

One possible choice is to be a plagiarist: $T$ is the set of all texts that already have been written, have a certain scientific quality, and/or can be found by searching the internet. Then operations like $f(a)\odot f(b)$ or $f(a)\oplus f(b)$ are  search queries to the internet. What have people already written about texts $a$ and $b$? What part of these writings describe differences or similarities of the texts? Starting with those found parts of texts, further operations $\oplus$ and $\odot$ just reduce the provided sets of search results. 

Another possible choice is to be a super-specialized linguist. Then $T$ is a set of very specific and obvious statements like ``The author uses/does not use a certain grammatical form''. Such findings about a text can be performed by a machine (see e.g. \cite{Biber1993}). However, restrictions like this concentrate the impact and expressiveness of our automated comparative text analysis on just linguistic features, leaving the literary analysis aside.

\subsection{Stone's representation theorem}\label{sec:stone}

The complicated part of text analysis seems to be the mapping $f$ from the set of possible texts $S$ to the power set ${\cal P}(S)$ of possible texts about texts. By restricting $S$ to a finite set of ``essays'' $T$, there is a way to represent the power set of $T$ in a computer.  Assuming an element $A$ of ${\cal B}'={\cal P}(T)$, i.e. $A$ is a subset of $T$, and let us further assume that $T$ has $n$ elements, then for each element of $T$ we just have to decide whether it belongs to $A$ or not. The representation of the $2^n$ different elements of $\cal B'$ is given by an $n$-digit binary number. Every digit just decides whether the corresponding element of $T$ belongs to $A$ or not. 

As an example let $T$ consist of four elements 
\begin{itemize}
    \item "The author does not use metaphors",
    \item "This is an English text",
    \item "There is a happy end", and
    \item "This novel will inspire many readers".
\end{itemize}

Now, take two texts $a$ and $b$ and check, whether the statements hold or not. A possible representation could be:  $A=f(a)=(0,1,1,0)$ and $B=f(b)=(1,1,0,0)$. If $T$ consists only of very simple statements that can easily be checked, then it would be possible to construct a machine, which provides this $n$-digit binary number for a given input text. This would be our mapping $f$. $f$ transforms a given text to an $n$-digit binary number.

Stone's representation theorem \cite{stone} is a generalization of this kind of binary number representation for {\em infinite} Boolean rings. Since $T$ is a finite subset, our situation is much less complex. 

Coming back to the example $A=(0,1,1,0)$ and $B=(1,1,0,0)$. Given two $n$-digit numbers, we have to explain the operations $\odot$ and $\oplus$. They are just ``bit''-wise operations on the binary numbers. For $\odot$ we have to perform the Boolean $\land$-operation (``and'') on each digit. For $\oplus$ we have to perform the Boolean $\dot{\lor}$-operation (``exclusive or'') on each digit. These operations are defined in Table~\ref{tab:my_label}. The result is again a binary number which represents an element of $\cal B'$. In the above example $A\oplus B=(1,0,1,0)$ and $A\odot B=(0,1,0,0)$.

\begin{table}[ht]
    \centering
    \begin{tabular}{c||c|c}
         $\land$ & 0 & 1 \\
        \hline
        \hline
         0 & 0 & 0 \\
         \hline
         1 & 0 & 1
    \end{tabular}
    \quad
        \begin{tabular}{c||c|c}
         $\dot{\lor}$ & 0 & 1 \\
        \hline
        \hline
         0 & 0 & 1 \\
         \hline
         1 & 1 & 0
    \end{tabular}

    \caption{Boolean operations carried out on every digit.}

    \label{tab:my_label}
\end{table}

It seems that the computations of $\odot$ and $\oplus$ are easy, whereas the complicated part is only the restriction of $S$ to a suitable subset $T$. If we want to do explicit restrictions and compter-based comparative text analysis, then we need to select a subset $T$ such that the mapping $\phi(f(\cdot)):S\rightarrow \{0, \ldots, 2^n\}$ is easy to be carried out. This is the same problem like in numerical mathematics, where we have to find a suitable representation of real numbers. In automated comparative text analysis, we have to find a suitable way to "represent" the properties of given texts in the computer.  

\subsection{Efficient representation of ${\cal B'}$} \label{sec:central}

Some quantitative linguistic approaches explicitly make use of representing texts by pre-defined vectors. If we, e.g., count the number of occurrences of certain grammatical or lexical constructions inside a text and write down these different numbers, then we end up with a vector of (natural, i.e.) real numbers representing the text. After this "coding" we then implicitly  assume, that everything which can be said about the text is represented by this vector ("everything which can be said" is exactly the role of $\cal B$).

Our approach is based on Stone's representation theorem. In principle it says that $\cal B$ is not represented by a continuous connected space. However, there is a way to end up with discrete points in a vector space representing all properties of texts in our approach, too.  After restricting from $S$ to a subset $T$ with $n$ elements, we can represent the properties of texts with an $n$-digits binary number. These numbers can also be regarded as (special) discrete points in an $n$-dimensional vector space.  However, there might be a more efficient representation. There are logical relations between texts and possible essays about texts. Not every point $\{0,1\}^n$ in the $n$-dimensional space is a possible representative of a text. There must be (complex) logical restrictions. The input texts may be regarded as elements of a sub-manifold in the $n$-dimensional space, such that (by an embedding theorem, e.g. \cite{embedding}) it can be mapped to points in an $m$-dimensional real vector space with hopefully $m\ll n$. A very similar idea is used in natural language processing by {\em word embedding} or {\em thought embedding} \cite{wordem}. After this embedding, the entries of the vector are not $0$ and $1$ anymore. $\phi(f(\cdot))$ is then a mapping from $S$ to discrete points in $\R^m$. 

Classical quantitative text analysis as well as our approach end up with an embedding of texts into an $m$-dimensional vector space. The question is now, how do we proceed with these vectors? In our approach: If we accept the existence of a complicated mapping $\phi(f(\cdot))$ from $S$ to a complicated manifold, then the computations of $\oplus$ and $\odot$ are not obvious ``bit''-wise operations anymore on this manifold. They are complicated, too. We will have to build machines to do these operations.

In the classical approach: Computer-based operations on vector spaces are mostly continuous operations on continuous spaces. Can we assume that our $\oplus$ and $\odot$ operations are represented by continuous functions? If we think of a continuous mapping between manifolds, then this question is connected to the existence of a continuous version of the operations in Table~\ref{tab:my_label}. This is also connected to the question of stability of the approach. There have been attempts in fuzzy logic to turn the operations into continuous operations. They can be constructed by $T$-norms \cite{Tnorms}. Daniel Greenhoe analyzed possibilities for a fuzzy Boolean ring based on theoretical results from 1970ies and 1980ies \cite{fuzzy}. Idempotency is only possible, if we define for the $n$ ``soft'' entries $a,b\in [0,1]$: 
$$a\land b:=\min\{a, b\} \text{ and } a\dot{\lor}b := \min\{\max\{a,b\},\max\{1-a,1-b\}\}.$$  However, this violates some of the properties of Boolean rings, like the existence of additive inverse elements for all continuous input values. Classical continuous approaches (including statistical analyzes) do not fit into the concept of Boolean rings. 

Our assumption is that a finite set of real-valued vectors are able to represent the elements of $\cal B'$. However, instead of defining in advance the rules to retrieve the components of these vectors for given texts, we incorporate the construction of the mapping $\phi(f(\cdot))$ into a construction of a more complex ``machine''.

We will not create the mapping $\phi(f(\cdot))$ by deeply thinking about representations of texts. Instead, we will focus on reducing the non-functionality of  ``continuous $\oplus$ and $\odot$''-machines on $m$-dimensional vector spaces. 

This change of paradigm can be justified by trying to exactly determine the point where complexity reduction takes place in TA. The complexity reduction of text analysis is not independent from the research question. In order to ask an interesting research question for a certain text, we need to compare this text to other texts. Thus, the complexity reduction of TA is based on comparing texts. We learn from errors and failing in these comparisons.

\subsection{Implicit restrictions}\label{sec:implicit}

Besides the obvious explicit choices of $T$, there is also a kind of ``black box'' approach which could be imaginable with the currently available computing technology, if we assume that the elements of $\cal B'$ can somehow be represented with real-valued vectors. 

Maybe there exists a machine (an artificial intelligence) which transforms a given text $a$ into a vector of real numbers. This machine represents somehow the nested mapping $\phi(f(a))$ with $\phi(f(\cdot)):S\rightarrow {\cal B}'$. The given text $a$ is mapped onto a real-valued vector $\phi(f(a))$. This is done without knowing how the real-valued vector can be identified with a subset of $T$ and even without explicitly knowing, how $T$ looks like. At this stage, it is not yet clear how to train this machine, but we will come back to that point later. 

For performing products and sums, we further construct two machines (again based on artificial intelligence). Such a kind of machine takes on input two real-valued vectors and its output is one real-valued vector. These machines represent the binary operations $\odot:{\cal B}'\times {\cal B}'\rightarrow {\cal B}'$ and  $\oplus:{\cal B}'\times {\cal B}'\rightarrow {\cal B}'$. Since we do not know the correspondences between the vectors and the elements of $\cal B'$, these two operations are not ``bit''-wise anymore. However, the training of these two machines can be based on respecting the algebraic properties of the operations ($\odot$  and $\oplus$). They can also be trained on the final output. This ``final output'' will be explained soon.

The last artificial intelligence machine which we have to construct can be named ``select a suitable representative of $\cal B'$''. This machine takes on input a real valued vector and selects on output one possible ``human readable'' text. This machine represents a mapping $\rho:{\cal B}'\rightarrow T$. Given a subset $X\in{\cal B}'$ of $T$, the machine selects one element $x=\rho(X)\in X$ of this set $X$. 

How to train these four artificial intelligences? If these machines work well, then the nested application of the machines of the form
\begin{equation}\label{eq:mach_x}
    x=\rho\Big(\phi\big(f(a)\big)\odot \phi\big(f(b)\big)\Big)
\end{equation}
would provide on output a text $x$ that includes statements which are valid for the input text $a$ as well as for the input  text $b$, whereas the nested application of the machines of the form
\begin{equation}\label{eq:mach_y}
    y=\rho\Big(\phi\big(f(a)\big)\odot\big(\phi\big(f(a)\big)\oplus \phi\big(f(b)\big)\big)\Big)
\end{equation}
would always provide on output a text $y$ that includes statements which are valid for the input text $a$ but not for input text $b$. These two compositions (and further compositions can be invented) are useful to train the four machines. If this training is successful, then we will posses machines for computing $\odot$ and $\oplus$. The advantage of this strategy is to be able to extract these two machines for $\oplus$ and $\odot$ and combine them into different networks in order to yield algorithms.

\subsection{An illustrative example}

In order to illustrate how the described four machines are combined, here is an illustrative example. We selected five different fairy tales in the version of the Grimm brothers. 
\begin{itemize}
\item[m1=] $\phi(f($Sleeping Beauty$))$ 
\item[m2=] $\phi(f($Snow White$))$ 
\item[m3=] $\phi(f($Cinderella$))$
\item[m4=] $\phi(f($Hans in Luck$))$
\item[m5=] $\phi(f($The Wolf and the Seven Little Goats$))$
\end{itemize}

In this assignment $\phi$ denotes the restriction of the possible statements about these fairy tales to a subset $T$, which should be unknown right now. Thus, we end up with elements $m1,\ldots, m5\in{\cal B'}$. 
For testing the following commands inside Octave\textregistered  \, one can use the software in Appendix~\ref{app:softw}.

\paragraph{Simple calculations.} In Sec.~\ref{sec:implicit}, two different compositions of the four machines are shown. In order to illustrate a possible result of the composition $m1\odot m2$ in (\ref{eq:mach_x}), the following commands can be used:
\begin{verbatim}
rho(m1 * m2)
ans = The fairy tale ends with a wedding.
\end{verbatim}
The multiplication $m1\odot m2$ leads to the set of all possible statements which are possible for ``Sleeping Beauty'' as well as for ``Snow White''. This special subset of $T$ is an element of $\cal B'$. From this subset, the function $\rho(m1\odot m2)$ selects one element: ``The fairy tale ends with a wedding''. 

In order to illustrate a possible result of the composition $m1\odot(m3\oplus m1)$ in (\ref{eq:mach_y}), the following commands can be used:
\begin{verbatim}
rho(m1 * (m3 + m1))
ans = The main character is noble by birth.
\end{verbatim}
The term $m1 \odot (m3\oplus m1)$ leads to the set of all possible statements which are possible for ``Sleeping Beauty'' but not for ``Cinderella''. From this subset, the function $\rho(m1\odot (m3\oplus m1))$ selects one element: ``The main character is noble by birth''.

\paragraph{Characteristics of clusters.} The determination of a statement which differentiates between two clusters of fairy tales, like in Sec.~\ref{sec:LR}, can also be illustrated in this way. In order to determine a statement which is possible for (the left stack) ``Sleeping Beauty'', ``Snow White'', ``Cinderella'', and ``Hans in Luck'', but not for (the right stack) ``The Wolf and the Seven Little Goats'', is given by
\begin{verbatim}
L=m1*m2*m3*m4;
R=(one+m5);
rho(L*R)
ans = The main character is a human.
\end{verbatim}
The set of possible statements characterizing the left stack is computed by the product $L\odot R$. One possible statement is: ``The main character is a human''. Another example:
\begin{verbatim}
L=m2*m3;
R=(one+m4)*(one+m5)*(one+m1);
rho(L*R)
ans = A "wicked stepmother".
\end{verbatim}

\paragraph{Embedding idea.} In Sec.~\ref{sec:central} it has been conjectured that the elements of $\cal B'$ can be represented efficiently by $m$-dimensional real vectors. In the illustrated software, the elements of $\cal B'$ are represented by $(m=1)$-dimensional ``vectors'':
\begin{verbatim}
m1.value
ans =  0.026341
m2.value
ans =  0.031486
\end{verbatim}
Other vales are {\bf m3=0.015113, m4=0.0080321, and m5=0.0010005}. The operations $\odot$ and $\oplus$ are carried out on the corresponding vector space:
\begin{verbatim}
(m1+m2).value
ans =  0.0050125
\end{verbatim}
Also the elements $0,1\in {\cal B'}$ are ``somewhere'' in this space ($'0'=0.00$ and $'1'=0.031486$).  In order to calculate the multiplicative neutral element $1$ of this restricted ring with regard to those statements which are important for the selected input texts, the following operations have to be carried out (more explanations about this will follow in Sec.~\ref{sec:kernel}):
\begin{eqnarray*}
    a &=& m1\oplus m2 \oplus (m1 \odot m2)\cr
    b &=& a\oplus m3 \oplus (a \odot m3)\cr
    c &=& b\oplus m4 \oplus (b \odot m4)\cr
    \mathrm{one} &=& c\oplus m5 \oplus (c \odot m5).
\end{eqnarray*}
The additive neutral element $0$ can be computed by \begin{equation*}
  \mathrm{zero}=m1\oplus m1.
\end{equation*}
The equality $m2.value=one.value$ means that statements which are not valid for ``Snow White'' are not included in $T$. For example, the computation of a statement which is valid for ``Sleeping Beauty'' but not for ``Snow White'' 
\begin{verbatim}
rho(m1*(one+m2))
ans = NULL
\end{verbatim}
results in an empty set. In order to include such statements, the four machines have to be retrained. The question of ``Efficient representation of $\cal B'$'' is now: Whenever a possible composition of our four trained machines leads to an unexpected result, which machine has to be replaced? How to correct for this ``error'' by keeping other  ``correct'' results mostly unaffected? A kind of task that our own brain does permanently. 

\paragraph{Training of $\phi(f(\cdot))$.} The real values corresponding to the five fairy tales are mentioned above. Now, we want to add a further tale to this set of tales: Rapunzel. The task of the mapping $\phi(f(\cdot))$ is to find the corresponding value which represents this fairy tale. If we know the set $T$ then it would be possible to select the corresponding statements and compute this value. However, there is also another way of finding this number. We asked a friend (having a natural neural network in his brain) to look at the five fairy tales and at the given values. Without knowing our criteria (that's the central point!), we asked him to figure out a suitable value for the fairy tale ``Rapunzel''. This means, he had to figure out, what the relation between the values and the content of the texts could be and to guess an own number. He guessed the value $0.010356$. The ``perfect'' value would have been $0.010050$. However, we did not find any composition of $\oplus$ and $\odot$ operations which would lead to a wrong result for $m6.value=0.010356$. It would be interesting to construct an artificial neural network with the same ability to find a perfect value (which would be TA). We have seen that the quality assessment of this mapping is not independent from the other values and from the intended comparisons, i.e., machine compositions (which would be CTA). 

\paragraph{One concrete preliminary architecture of an artificial neural network.} Fig.~\ref{arch} shows how the machines can be realized in a computer. If we have in mind that a software code like that in the Appendix~\ref{app:softw} can be used to implement the calculations of $\oplus$ and $\odot$, then these operations are not to be learned but already available.  Then training of the other two machines $\phi$ and $\rho$ can be done like following: 
\begin{figure}[ht]
\centering
\includegraphics[width=0.95\textwidth]{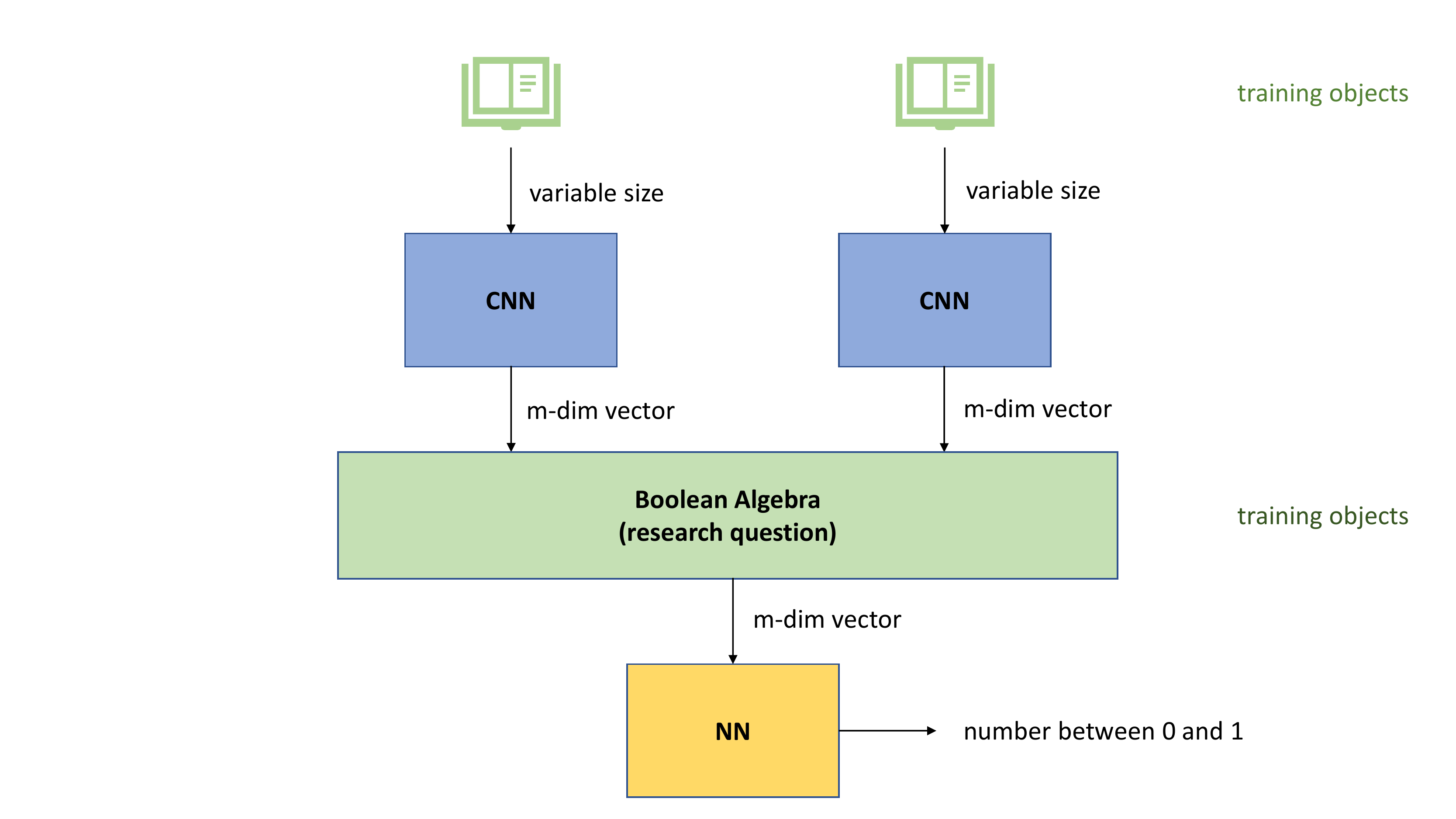}
\caption{\label{arch} A possible architecture to realize the machines by artificial neural networks.}
\end{figure}
We take an arbitrary text resource (this is our $T$). A number between $0$ and $1$ which is the output of $\rho$ points to a relative position within this text $T$. The sentences which can be read at this position is the result of our machines. Therefore, the output of $\rho$ is denoted as a number in Fig.~\ref{arch}. $\rho$ is represented by a orange box - an artificial neural network to be trained. On input we have two times the same(!) neural network representing $\phi$ (blue boxes). This network should be a convolutional neural network, because the input is a text of unknown length. This network turns text into a real-valued vector. Also this CNN has to be trained. What is the training data? Besides the different available input texts, the training data also consists of different possible compositions of $\odot$ and $\oplus$. The training is, thus, not only based on the given texts but also on the research question of comparative text analysis represented by the composition of the algebraic operations. 

\subsection{The algebraic structure of $\cal B'$}

Besides all possible difficulties in constructing machines of the described types, there is always a structural problem with restricting $S$ to a subset $T$. The problem is given by the kernel of the mapping $\phi$. This kernel $\mathrm{ker}(\phi)$ is the set of all elements in $\cal B$ which are mapped via $\phi$ to zero $0 \in\cal B'$. {All the statements about the texts which we decided {\em not} to write down lead to a bias in our comparative text analysis.} From algebra (theory of ring homomorphisms) it is known, that $\mathrm{ker}(\phi)$ is an ideal in $\cal B$ and that $\cal B'$ is isomorphic to ${\cal B}$\textbackslash$\mathrm{ker}(\phi)$ (cf. \cite{bosch}). $\mathrm{ker}(\phi)$ is even a principle ideal generated by the complement of $T$. 
\begin{definition} (cf. \cite{bosch})
An {\em ideal} $I$ of a commutative ring ${\cal B}$ is a subset $I\subseteq {\cal B}$ such that
\begin{itemize}
    \item[(i)] $0\in I$,
    \item[(ii)] for all $a,b\in I$ also the sum $a\oplus b\in I$,
    \item[(iii)] for all $a\in I$ and $b\in {\cal B}$ it holds $a\odot b\in I$.
\end{itemize}
The ideal $I$ is said to be generated by the elements $a_1,\ldots, a_n\in{\cal B}$ denoted as $I=\langle a_1,\ldots,a_n\rangle$, if it is the intersection of all ideals of $\cal B$ which include these elements.
A {\em principle ideal} is an ideal $\langle a\rangle$ which is generated by only one element $a\in{\cal B}$. A {\em principle ideal ring} is a ring in which every ideal is a principle ideal. 
\end{definition}
Zero-divisors are non-zero elements $A,B$ of a ring with $A\odot B=0$. The real numbers do not include such zero-divisors. The real numbers are also an example for an algebraic field, because for every non-zero real number $r$ there exists an inverse of multiplication $1/r$. Although, it might not be clear yet, how the above representation of $\cal B'$ can be used in quantifying the bias of our machine, it shows that there is a deep algebraic structure in restricting our point of view to certain aspects of text analysis. The question, whether we are able to select $\cal B'$ in such a way, that it turns into something, that can be handled easily with the help of machines, is an algebraic question.

Algebra studies the question of extending the ideal $\mathrm{ker}(\phi)$ by reducing the set $T$ is such a way, that the ring $\cal B'$ does not have any zero-divisors or such that it even turns into an algebraic field. The possibility of restricting $T$ accordingly is connected to the question, whether an ideal in $\cal B$ can be extended to a prime ideal.  For this question, mathematicians have already given positive answers \cite{maximalId}. However, the theory of prime ideals and ultrafilters is not yet of practical use for comparative text analysis. 

Maybe, the existence of zero-divisors is even the most important structural property of Boolean rings. This property will also be used for many implications in the followings.

\section{Clustering texts}

Imagine now a set of texts (again the mappings $f$ of these texts are meant). If these texts are assigned to two different stacks, then it might be ``easy'' to figure out, what is the difference between these two stacks. This means, it might be easy to compute $L\odot R$ mentioned in Sec.~\ref{sec:LR}. {\em Clustering} is more than that. If someone tells us that there is a difference between the given two stacks of texts, then we will find differences. In solving the clustering problem, we are not provided with the stack assignment in advance. Comparative texts analysis intends to substantiate a possible hypothesis (Ex1 in Sec.~\ref{sec:LR}) which depends on not knowing the stack assignment in advance. Comparative text analysis also has to find the assignment in order to reveal something meaningful (Ex2). The combinatorial problem of CTA is to assign given texts to two different stacks LEFT and RIGHT, such that $L\odot R$ is optimal. But what does it mean to be optimal?

\subsection{Partial ordering of $\cal B$}

Checking for optimality in real-world applications is often connected with the question of finding a situation in which a certain assessment value (quantifying the ``quality'' of that situation) is minimal or maximal. At a first glance, texts can not be sorted according to ``minimal'' or ``maximal'' if we also want to consider their quality or inherent richness. However, there is a partial ordering of the set $\cal B$. $A\in{\cal B}$ is said to be greater than $B\in{\cal B}$, i.e., $A\geq B$, if and only if the following relation holds $B\odot (A\oplus B)=0$. It can be shown easily with the above properties of a Boolean ring, that for all $A,B,C\in{\cal B}$
\begin{itemize}
    \item $A\geq A$ (reflexivity),
    \item that $A=B\Leftrightarrow A\geq B\wedge B\geq A$ (antisymmetry),
    \item and that $A\geq B\wedge B\geq C\Rightarrow A\geq C$ (transitivity).
\end{itemize}  
Easy to show are also the relations $A\geq 0$ and $1\geq A$ for all elements $A\in{\cal B}$ and $C\odot A\geq C\odot B$ if $A\geq B$. Interestingly, $A\geq A\odot C$ for all $A,C\in{\cal B}$. Thus, there is indeed an extreme situation $L\odot R=0$, which will be discussed in Sec.~\ref{sec:NP}. 

The product $L\odot R$ depends on the assignment $\pi$ of the texts with regard to the two stacks. This will be expressed via $(L\odot R) (\pi)$. The situation $(L\odot R)(\pi)=0$ means that common properties of stack LEFT are also part of the elements on stack RIGHT, which is usually not the intended target of a text clustering based on comparative text analysis. Maybe the opposite target is interesting. Find an assignment $\bar{\pi}$ of given texts to two stacks LEFT and RIGHT, such that $C=(L\odot R) (\bar{\pi})$ has the following property. For all possible assignments $\pi$ it holds $(L\odot R)(\pi)\not> C$. This means, that there does not exist an assignment $\pi^\ast$, such that the product $(L\odot R)(\pi^\ast)\not=C$ is greater than $C$. However, having the size of $\cal B$ in mind, this criterion is probably met, if all texts go to stack RIGHT. Or it might even be possible, that the products $L\odot R$ are not comparable with this relation $\geq$, which is also very likely. Then every assignment is optimal. The problem is not to compute $L\odot R$, the main problem is to {\em find} the suitable assignment $\pi$, or a suitable ``ordering'', or a suitable permutation of the elements $X_1,\ldots,X_n$.
  
Maybe we search for alternative clustering approaches in mathematics  to compute this ordering. Usually we define a clustering on vector spaces.  In terms of rings, the correspondent concept of a vector space over a field is denoted as {\em a module over a ring}.

\subsection{The module ${\cal B}^n$}

In order to transfer vector space concepts to our Boolean ring situation. We use the following  \begin{definition} (cf. \cite{bosch})
Given a commutative ring $\cal B$ with unity. A {\em module $M$ over this ring $\cal B$} is defined by the following properties. For all elements $u,v,w\in M$, for $\lambda,\mu\in{\cal B}$, and for two operations $\oplus:M\times M\rightarrow M$ and $\odot:{\cal B}\times M\rightarrow M$:
\begin{itemize}
\item[(i)] $(M,\oplus)$ is an Abelian group, i.e., 
    \begin{itemize}
        \item[a)] there exists an element $0\in M$ such that $0\oplus u=u$, 
        \item[b)] for every $u\in M$ there is an $\bar{u}\in M$ with $u\oplus \bar{u}=0$, 
        \item[c)] $u\oplus v=v\oplus u$, and $u\oplus(v\oplus w)=(u\oplus v)\oplus w$. 
    \end{itemize}
\item[(ii)] $1\odot u=u$, 
\item[(iii)]$\lambda\odot(u\oplus v)=(\lambda\odot u)\oplus (\lambda\odot v)$, and
\item[(iv)] $(\lambda\oplus \mu)\odot u=(\lambda\odot u)\oplus(\mu \odot u)$.
\end{itemize} 
A trivial choice of a modul over the ring $\cal B$ is given by the cartesian product ${\cal B}^n$ with element-wise sums and with element-wise multiplication of a ring element. An element $v\in {\cal B}^n$ of this module is denoted as {\em modus} (plural: {\em modi}). A {\em sub-module $N$ of $M$} is a module which is a subset of $M$.
\end{definition}

A modus $m\in{\cal B}^4$ can, e.g., be written down like this
$$
m=(a,c,b,d)^T \text{ or } m=\begin{pmatrix}a\\ b \\ c\\ d\end{pmatrix},
$$
where $a,b,c,d\in {\cal B}$. 
With these preparations, also matrices can be defined ${\cal M}\in {\cal B}^{n\times m}$. A matrix is a table-like scheme in which every element ${\cal M}_{ij}$ stems from the ring $\cal B$. $i=1,\ldots,n$ is the row index and $j=1,\ldots, m$ is the column index. Examples for matrices can be found in Sec.~\ref{sec:example}. The $i$-th component of a matrix-modus multiplication $v={\cal M}w$, where $v\in {\cal B}^n$ and $w\in{\cal B}^m$, is defined by
$$ 
  v_i = ({\cal M}_{i1} \odot w_1) \oplus \ldots \oplus ({\cal M}_{im} \odot w_m).
$$
Matrix-matrix products can be defined accordingly, but they are not used in the followings.

\begin{definition} (cf. \cite{bosch})
A family  $F:=\{b_{i}\mid i\in I\}$ of elements of a module $M$ over a ring $R$ is {\em free}, if for every finite subset $\textstyle J\subseteq I$ and for all $\textstyle r_{i}\in R$:
$$\sum _{i\in J}r_{i}\odot b_{i}=0\;\Rightarrow \;\forall i\in J\colon \,r_{i}=0,$$ 
where the $\sum$-symbol is used for $\oplus$-operations. The expression $\sum_{i\in J} r_i\odot b_i$ is denoted as {\em linear combination of modi}. By varying $r_i$ {\em the modi span/generate a sub-module $N$}. If this sub-module is equal to $M$, then the family $F$ is denoted as {\em basis of $M$}. The elements of a basis are denoted as {\em basis modi}. A module that has a basis is denoted as a {\em free module}. 
\end{definition}

A module does not need to have a basis. A free module can have different bases. These bases can have a different number of modi. Some modules have the property, that the basis has a fixed number of modi, such modules are denoted as {\em IBN-modules} (IBN = invariant basis number). ${\cal B}^n$ is a free IBN module. $n$ is the {\em dimension} or {\em rank} of this module.  

\subsection{Idea of spectral clustering}\label{sec:PCCA+}

With these preparations we can transfer the ideas of spectral clustering to the situation of Boolean rings. 

Vector spaces: In trying to cluster $n$ point vectors, spectral clustering is based on creating a matrix ${\cal M} \in{\R^{n\times n}}$ which represents a linear mapping ${\cal M}:\R^n \rightarrow \R^n$. The basic idea is to generate the element ${\cal M}_{ij}\in\R$ of this matrix by ``comparing'' the $i$-th and the $j$-th point vector. In real vector spaces it can be based on the Euclidean distance between the two point vectors or on similarity measures (or simply on their dot product). The heuristics is now, that columns of this matrix $\cal M$ are similar whenever the corresponding point vectors, which produce these columns, are similar. Thus, (dominant) eigenvectors will have similar entries in their components $i$ and $j$ if the $i$-th and the $j$-th point vector are similar. An ``ordering'' of the entries of these eigenvectors lead to the desired permutation which reveals the clustering -- also known as Fiedler's cut \cite{Fiedler}.   

Boolean rings: Transferred to the situation of Boolean rings, we have to compute a matrix ${\cal M}$ which represents a linear mapping in the module ${\cal B}^n$. The elements of this matrix stem from $\cal B$. If the element ${\cal M}_{ij}$ is supposed to depend on a comparison of $X_i\in{\cal B}$ and $X_j\in{\cal B}$, it can, e.g., be defined as ${\cal M}_{ij}=X_i\oplus X_j$ or as ${\cal M}_{ij}=X_i\odot X_j$. The matrix $\cal M$ may have eigenmodi $v\in{\cal B}^n$ now, such that ${\cal M}v=\lambda\odot v$ for some $\lambda \in{\cal B}$, but eigenvector computation and eigenvector theory in real spaces need not be transferable to rings. More general, the matrix $\cal M$ will have invariant sub-modules, i.e., for the matrix $\cal M$ there exist modules $E\subset {\cal B}$ with the following property: For every $v\in E$ it holds that ${\cal M}v\in E$.  

If we are able to compute these invariant sub-modules of $\cal M$, would this help us to solve the cluster problem? Even if we know the modi which span the sub-modules and even if we know their element-wise entries, the partial ordering defined on $\cal B$ might again be useless, because not all entries are comparable with the others. Furthermore, a ``one-dimensional'' invariant sub-module need not exist, which means that we would have to order multiple modi at once, which is not defined yet. However, these invariant sub-modules will have a structure which will depend on the similarity of the elements $X_i$. 

A special invariant sub-modul of a matrix is given by the kernel of the matrix $\cal M$, which will be exemplified in Sec.~\ref{sec:example}. Especially in the case of ${\cal M}_{ij}=X_i\odot X_j$ the kernel of $\cal M$ has a very special structure, which provides a possibility to extract the characteristics of the clusters from that kernel. This will also be exploited in the next section. However the theory will be explained for a ($2$-dimensional modi) Gramian matrix instead for the (one-dimensional modi Gramian) matrix $\cal M$.

\subsection{All possible clusterings: kernel of a Gramian matrix}\label{sec:kernel}

We will do the following thought experiment: Assume there is a set of elements of $X_1,\ldots, X_k\in\cal B$. Futher assume, that there is also a reasonable clustering of these elements into two subsets LEFT and RIGHT, such that there exist properties which are ``exclusive'' only for one of the two clusters. The problem is, that we do not know the assignment of the elements to the clusters. Now, take $n$ pairs of the $k$ elements selected from $\cal B$ leading to $n$ different two-dimensional modi of the form $(X_i,X_j)$. There are four different possibilities when thinking of the unknown cluster assignment, schematically:  $(L,L), (L,R), (R,L),$ and $(R,R)$. 

The next step is to compute the $n\times n$-Gramian matrix $\cal G$ for the $n$ different pairs, where the {\em dot product} is defined as $$(X_a, X_b)\cdot(X_c,X_d)=(X_a\odot X_c) \oplus (X_b\odot X_d),$$ i.e., the element ${\cal G}_{st}$ of the Gramian matrix is the dot product of the $s$-th and the $t$-th pair.  

We assumed the clustering LEFT and RIGHT to be ``reasonable''. This means that there should exist non-zero elements $l,r\in {\cal B}$ such that $l\odot X_i=0$ for all elements $X_i\in$ RIGHT and $l\odot X_i=l$ for all elements $X_i\in$ LEFT. Furthermore, $r\odot X_i=0$ for all elements $X_i\in$ LEFT, whereas $r\odot X_i=r$ for all elements $X_i\in$ RIGHT. With these two ring elements we can construct a non-zero modus $v\in{\cal B}^n$, such that ${\cal G}v=0$. Whenever the $i$-th pair is of the form $(R,R)$, we set $v_i=l$, and whenever the $i$-th pair is of the form $(L,L)$, then the entry is $v_i=r$. All other entries of $v$ are zero. 

This modus $v$ is an element of the kernel of $\cal G$.
$$\mathrm{ker}({\cal G}):=\{v\in {\cal B}^n; {\cal G}v=0\}.$$
Since we do not know the assignment of the $X_i$ to the two clusters, we can not construct $v$ in advance. However, we can maybe compute the kernel of $\cal G$ and then we know that $v$ must be an element of $\mathrm{ker}({\cal G})$. For {\em every} reasonable clustering of the elements $X_i$, we will find a corresponding modus in the kernel of $\cal G$. Thus, the kernel of ${\cal G}$ includes {\em all} reasonable $2$-clusterings of the texts. 

The kernel of a matrix is a sub-module of ${\cal B}^n$.
The module ${\cal B}^n$ is free. 
This means that there is a clear restriction of the interpretation margin in comparative text analysis, when thinking of assigning given texts to two different clusters. 

For the matrix $\cal G$ we need an algorithm (with algebraic operations of the form $\oplus$ and $\odot$) to compute its kernel. Is there a way to do it? 

\paragraph{Method 1 for kernel computation.} If we have in mind to restrict our Boolean ring ${\cal B}$ to a {\em finite} Boolean ring ${\cal B}'$, then every ideal of this finite Boolean ring is trivially finitely generated. Hence, since every finitely generated ideal in a Boolean ring is principal $\langle a_1,\ldots,a_r\rangle =\langle a \rangle$ (where $a$ is just the ``union'' of the elements $a_1,\ldots,a_r$), the Boolean ring $\cal B'$ is a principal ideal ring \cite{mathstack, mathstack2}.

$({\cal B}')^n$ is a free module and $\mathrm{ker}({\cal G})$ is a sub-module. Is this sub-module free? The answer is yes, a basis exists. Here is an example of how to compute the basis of sub-modules of free modules over commutative principal ideal rings \cite{basismodule}:

$M=({\cal B'})^n$ is a free module over the principal ideal ring $\cal B'$ with basis $m_1,\ldots m_n$. $N=\mathrm{ker}({\cal G})$ is a sub-module of $M$. A basis of $N$ can be computed iteratively: Define $N_{i}=N\cap \langle m_{1},\dotsc ,m_{i}\rangle$. Let the ideal 
$$\{r\in {\cal B'}:\exists m\in N_{{i+1}}{\text{ with }}m=m'\oplus (r\odot m_{{i+1}}){\text{ and }}m'\in \langle m_{1},\dotsc ,m_{i}\rangle \}$$
be generated by $a_{{i+1}}\in{\cal B'}$. Furthermore, $n_{{i+1}}=m'\oplus (a_{{i+1}}\odot m_{{i+1}})\in N_{{i+1}}$ with  $m'\in \langle m_{1},\dotsc ,m_{i}\rangle$, then the basis of $N$ is given by the direct sum of $N_1$ and of all different subrings $n_{i}\odot {\cal B'}$ that have been found by this algorithm.    

\paragraph{Method 2 for kernel computation.} A more practical way to find elements of the kernel of $\cal G$ starts with a random modus $v\in ({\cal B}')^n$. Applying the matrix $\cal G$ leads to a modus $w={\cal G}v$ which is not equal to the zero modus in general. For this modus we compute a ring element $\lambda\in\cal B'$ via $\lambda =(1\oplus w_1)\odot \ldots \odot (1\oplus w_n)$. The ring-modus-product $\lambda\odot w$ is zero. Thus, $x=\lambda\odot v$ is an element of the kernel of $\cal G$, which can be shown via ${\cal G}x={\cal G}(\lambda\odot v)=\lambda\odot ({\cal G}v)=\lambda\odot w=0$. Another possibility is to find two matrices ${\cal G}_1, {\cal G}_2$ such that ${\cal G}_1\oplus {\cal G}_2={\cal G}$. Again with a random modus $v$ we compute ${\cal G}_1v=u$ and ${\cal G}_2v=w$. Now the ring element is computed via $\lambda=(1\oplus u_1\oplus w_2)\odot\ldots\odot (1\oplus u_n\oplus w_n)$, with $\lambda\odot u=\lambda \odot w$. In this case, $x=\lambda\odot v$ is an element of the kernel of $\cal G$. Having in mind the algorithmic details of Section \ref{sec:implicit}, we can use the trained machines to carry out these types of operations. The only problem remaining is to determine the element $1$ on the $m$-dimensional manifold. The element $1$ is the ``union'' of all elements in $T$ or of a (maybe overlapping) covering of $T$ by subsets $A_1,\ldots,A_k\in {\cal B'}$. This union can be constructed with the aid of an iteration. Note, that the union of two elements $A_1,A_2$ of $\cal B'$ is given by: $A_1\oplus A_2 \oplus (A_1\odot A_2)$.   

\paragraph{How to proceed?} Given a basis of the kernel of $\cal G$, an idea of Robust Perron Cluster Analysis may help to interpret the result \cite{weberPhd}. Assume we have found $r$ modi which span $\mathrm{ker}(\cal G)$, then there might exist a matrix ${\cal A}\in {\cal B}^{r\times r}$ (a basis transformation matrix) such that applying this matrix to these $r$ modi leads to another set of modi $\chi_1,\ldots,\chi_r$, which then also span a sub-module of the kernel of $\cal G$. How does an optimal transformation matrix look like?

Note, that using Gramian matrices for clustering is advantageous, because after applying $\cal A$ some elements $i$ of the kernel basis modi will become zero. If the index $i$ belongs to columns of the matrix created by $(L,R)$ or $(R,L)$, then the corresponding kernel basis modus (indicating this $LR$-clustering) should be zero in that component. Only components which belong to $(R,R)$ or $(L,L)$ are non-zero. Therefore, transformations $\cal A$ which lead to feasible sparsity pattern in the basis modi are searched for. In this case an ordering of elements of the kernel modi is not needed.  This will be exemplified next. 

\subsection{An example}\label{sec:example}

We will show a very simplified  example based on a subset $T$ of texts which includes only four elements, like in Sec.~\ref{sec:stone} with the corresponding binary number representation. Note, that using a vector-valued representation of texts would be the "natural representation" when applying computers and machine learning to train the four mappings. The simplified situation is only for illustration. Furthermore, we want to cluster the four elements $$X_1=1100, X_2=0111, X_3=1001, \text{ and } X_4=0011.$$ We search for a reasonable clustering like in Sec.~\ref{sec:kernel}. Thus, we search for two non-zero zero divisors $r,l\in\cal B'$ with $l\odot r=0$.  The four elements should be grouped such that every multiplication of an element of LEFT with $l$ leads to $X\odot l=l$ and every multiplication with $r$ is equal to zero. In group RIGHT every multiplication with $l$ is zero and every multiplication with $r$ is $r$. 

\paragraph{The simple way.} In the situation of a finite set $T$ and, thus, a finite ring $\cal B'$, there is an algorithm to find $l$ and $r$. We simply multiply $X_i$ with {\em all} non-zero elements $a$ of $\cal B'$. A suitable multiplication (in this case: $a=1010$) directly shows the clustering: $X_1\odot a=X_3\odot a=1000$ and $X_2\odot a=X_4\odot a=0010$. Thus, $l=1000$ and $r=0010$ and LEFT$=\{X_1,X_3\}$, RIGHT=$\{X_2,X_4\}$. However, imagine $T$ to be the set of all internet texts. Or imagine the situation of Sec.~\ref{sec:implicit} where $T$ is not known explicitly, then multiplication with all elements of $\cal B'$ is computational impossible. 

\paragraph{The $\cal M$-matrix way.} We compute the matrix ${\cal M}_{ij}=X_i\odot X_j$:
$$
{\cal M}=
\begin{pmatrix}
1100 & 0100 & 1000 & 0000 \\
0100 & 0111 & 0001 & 0011 \\
1000 & 0001 & 1001 & 0001 \\
0000 & 0011 & 0001 & 0011 
\end{pmatrix}.
$$
One element of the kernel of this matrix is $m=(0111, 1100, 0111, 1101)^T$. This kernel modus is not very helpful, because it does not include two different elements $l$ and $r$ which have the further property $l\odot r=0$. The interesting thing is, that there exists an ordering of the elements of $m$ within the two different clusters. $0111 \geq 0111$ and $1101\geq 1100$. Whereas, this ordering does not exist between the two clusters: $0111$ can not be compared with $1100$ or with $1101$. In order to reveal the clustering, we need to know the ring element $a=1010$ which is multiplied with this kernel modus to provide $(0010, 1000,0010,1000)^T$, which is also a kernel modus of $\cal M$ and provides the desired $lr$-structure. 

\paragraph{The $\cal G$-matrix way.} Much more complex is the computation of the Gramian matrix $\cal G$ based on the pairs $\varphi_1=(X_1,X_2), \varphi_2=(X_2,X_3), \varphi_3=(X_1,X_4), \varphi_4=(X_1,X_3), \varphi_5=(X_2,X_4),$ and $\varphi_6=(X_3,X_4)$. Note, that this approach would not necessarily need to be based on all possible pairs. Given these six modi, the Gramian matrix is computed by the dot product  ${\cal G}_{ij}=\varphi_i\cdot \varphi_j$:
$$
{\cal G}=
\begin{pmatrix}
1011 & 1111 & 1011 & 1101 & 0111 & 1011\\
1111 & 1110 & 1011 & 1001 & 0010 & 1010 \\
1011 & 1011 & 1111 & 1101 & 0111 & 1010 \\
1101 & 1001 & 1101 & 0101 & 0101 & 1001 \\
0111 & 0010 & 0111 & 0101 & 0100 & 0010 \\
1011 & 1010 & 1010 & 1001 & 0010 & 1001
\end{pmatrix}.
$$
An element of the kernel of $\cal G$ is $m=(0010, 0010, 0000, 0110, 1100, 0100)^T$. First of all, this is not a helpful modus, because it has four different non-zero entries (instead of two). In order to get kernel modi which only have two different non-zero entries, we try a  multiplication with the complements of the four elements $0010, 0110, 1100, 0100$. This produces two $2$-valued modi ($e=1111$):
$$
\underbrace{(e \oplus 0010)}_{=1101}\odot m=
\begin{pmatrix}
0000\\
0000\\
0000\\
0100\\
1100\\
0100
\end{pmatrix},
\underbrace{(e \oplus 0100}_{=1011})\odot m=
\begin{pmatrix}
0010\\
0010\\
0000\\
0010\\
1000\\
0000
\end{pmatrix}.
$$
None of these modi fit to a feasible sparsity pattern. For example, take the first modus $(e \oplus 0010)\odot m$. The fifth and sixth entry is non-zero. These elements belong to the modi $\varphi_5=(X_2,X_4)$ and $\varphi_6=(X_3,X_4)$. This means that $X_2$ and $X_4$ belong to the same cluster as well as  $X_3$ and $X_4$ belong to the same cluster. From the transitivity rule, it follows that $X_3$ and $X_2$ belong to the same cluster, however the second element corresponding to $\phi_2=(X_3,X_2)$ is zero. Thus, the non-zero-pattern is not valid. It is the task of the clustering to find a feasible linear combination of kernel modi, such that the sparsity pattern fits to the clustering and that non-zero entries only have two different values. The following sum of kernel modi provides a feasible solution:
$$
(1011\odot m) \oplus 
\begin{pmatrix}
0010\\
0010\\
0000\\
0000\\
0000\\
0000
\end{pmatrix}=
\begin{pmatrix}
0000\\
0000\\
0000\\
0010\\
1000\\
0000
\end{pmatrix}.
$$
The sparsity pattern fits to the desired clustering LEFT$=\{X_1,X_3\},$ RIGHT$=\{X_2,X_4\}$. From the resulting modus one can read $l=1000$ and $r=0010$. A similar problem is solved in Robust Perron Cluster Analysis when trying to find a feasible transformation matrix $\cal A$ to create certain zero-entries in the vectors $\chi_i$ \cite{weberPhd}. 

\section{Complexity of the ${\cal B}$-language}\label{sec:complex1}

{Usually, natural language processing units which have already been realized on our computers are based on mappings $\theta:S\rightarrow S$. A given (spoken or written) text is transformed into another text, e.g., the text is translated from German into English. Some processing units convert (spoken) language into instructions that are used in order to steer a machine or to initiate a software. The instructions that are then sent to the computer are also a kind of language (machine language).} Text analysis is a mapping $f:S\rightarrow {\cal B}$. It can be seen as a kind of translation of texts into a new language - the ${\cal B}$-language. {\em Comparative} text analyis is based on a subset of texts and, thus, can be regarded as a transformation inside this $\cal B$-language, $\Theta:{\cal B}\rightarrow {\cal B}$. Note, that elements of $\cal B$ are just subsets of $S$. The complexity of this transformation is the objective of our studies.
 
\subsection{The $\cal B$-language}

We are not able to ``speak'' this $\cal B$-language. This language is very different from what we would call a natural language. An element $A\in {\cal B}$ is a subset of elements of $S$. Translating a text $a\in S$ into this language via $A=f(a)$ means that $A$ includes everything that can be written (maybe even thought, associated, felt..., till now and till eternity) in connection with $a$. Elements of $S$ are texts. Elements of $\cal B$ will be denoted as {\em pexts}, where the "p" is used to indicate this "power set" approach. 

If an author of ${\cal B}$-land writes a pext, then {\em everything} that could be said about this pext is already included in the pext. $\cal B$-land does not know about secondary literature. For instance, imagine a student in our world would get the text "Prometheus" and the task to do a comparative text analysis with ... "Prometheus". Probably the answer would be: "This is a stupid task, it's the same text twice!". In ${\cal B}$-land a student who gets the task to find out the commonalities of a pext $A$ and a pext $A$ will provide the pext $A$, because $A\odot A=A$. A pext $A$ does not provide any space for interpretation. 

The machines that have been constructed in Sec.~\ref{sec:implicit} learn to do comparative text analysis in the $\cal B$-language, however, with an extremely limited expressiveness $\cal B'$. After the first machine "$\phi(f(\cdot))$" translates the texts $a$ and $b$ into this $\cal B'$-language, comparative text analysis is carried out. This provides a pext in $\cal B'$-language that is back-translated via the fourth machine "$\rho$" into a text, that we can understand. 

However, there might be still a space for creativity. If we believe that transformations of a text $a$ into a text $\theta(a)$ in our world would also lead to a transformation of the pext $f(a)$ into a {\em different} pext $f(\theta(a))$, then inhabitants of $\cal B$-land have in fact the possiblity to be creative in a similar way we are. We should not think too post-modern about the limits and possibilities of writing texts about texts. Stricter limits in our world about the range of possible comparative text analyzes lead to more possible creativity in $\cal B$-land.  

\subsection{Transformations of the $\cal B'$-language}

In the moment that the machines $\odot$ and $\oplus$ have been trained perfectly, we invented and/or created a subset (a primitive dialect) $\cal B'$ of the very rich language $\cal B$. In fact, this procedure does not ``explain'' how to do comparative text analysis in our own language, but the machines $\odot$ and $\oplus$ {\em define} how to do it in $\cal B'$ without explicitly knowing the subset $T$. 

{Our own creativity is visible in being able to transform texts. We can add metaphors, write the negation of statements, translate texts into different natural languages, rearrange the sentences to emphasize different aspects by following the rules of grammar, and so on. This is very much in the spirit of Chomsky \cite{Chomsky}.   }

Let us assume, that the language $\cal B'$ is able to do similar non-trivial transformations $\Theta:{\cal B'}\rightarrow {\cal B}'$ based on their pexts. A pext $A$ is transformed into $\Theta(A)$. Can we somehow ``study'' the transformations $\Theta$? Given a certain transformation $\theta$ of texts, we are also able to define a corresponding transformation $\Theta$ of pexts as long as the input texts in $\cal B'$ are based on the mapping $\phi(f(\cdot)):S\rightarrow {\cal B}'$. $\Theta$ can be defined to be compatible with $\theta$ on this subset of pexts by making the following diagram commute:  
\begin{equation*}
    \begin{matrix}
    a &\longmapsto & \theta(a) \\    
    \downmapsto & &\downmapsto \\
    \phi(f(a)) &\longmapsto & \Theta(\phi(f(a)))=\phi(f(\theta(a)))
    \end{matrix}
\end{equation*}
The transformation $\Theta$ is such that $\Theta(\phi(f(a))):=\phi(f(\theta(a)))$. In order to be able to define $\Theta$ in such a way, it must be assured that texts $a$ and $b$ leading to the same pext $\phi(f(a))=\phi(f(b))$ also lead to the same transformations $\phi(f(\theta(a)))=\phi(f(\theta(b)))$.  This is true for injective mappings $\phi(f(\cdot))$. In $\cal B$ the mapping $f$ should be injective, however, $\phi(f(\cdot))$ could be non-injective for $\cal B'$.  The ``creativity'' of $\cal B'$ is visible through the complexity of its transformations $\Theta$. How to analyze this complexity? 

\subsection{Linearization of $\Theta$}

In order to analyze mappings between vector spaces in functional analysis, linearization of the transformation (i.e., the operator) is a standard tool. The Galerkin projection of operators play a crucial role. These projections lead to finite dimensional $m\times m$-matrices. Often a spectral analysis of such matrices is used (computing eigenvalues and eigenvectors) to characterize the operators.  

Assume we have a matrix $\cal T$ with entries from a Boolean ring $\cal B$. Then it is easy to show, that only the eigenvalues $1$ and $0$ play an important role. Take a modus $\varphi\in{\cal B}^m$ which is an eigenmodus of $\cal T$ with eigenvalue $\lambda\in{\cal B}$, then ${\cal T}\varphi = \lambda\odot\varphi$. Either $\lambda=0$ holds, or $\lambda \not= 0$ and $\lambda\odot\varphi\not= 0$ hold. In the latter case, multiplying the eigenequation with $\lambda$ leads to ${\cal T}(\lambda \odot\varphi)=\lambda\odot\lambda\odot\varphi=\lambda\odot\varphi$. Replacing $\lambda\odot\varphi\not= 0$ with $\xi$ leads to the equation ${\cal T}\xi=\xi$. This shows, that every eigenmodus can be restricted to an eigenmodus of eigenvalue $1$ or $0$. In this case, a ``generalized eigenvalue problem'' like ${\cal T}\varphi=\lambda\odot {\cal I}\varphi$ with two matrices $\cal T$ and $\cal I$ and $\lambda\in\{0,1\}$ can be solved by analyzing the kernel of ${\cal T}\oplus{\cal I}$.

In the setting of $\cal B'$, we can do a Galerkin-based approach of trial and test of a transformation $\Theta$ in the following way: First we need a set of pexts: $\varphi_1,\ldots,\varphi_m\in {\cal B'}$. These pexts are based on mappings of given trial texts of $S$ by applying the trained mapping $\phi(f(\cdot))$. Then we create an $m\times m$-matrix ${\cal T}$, where the element ${\cal T}_{ij}=\phi_i\odot \Theta(\phi_j)$. These expressions can be calculated on a computer simply by applying $\odot$ with the trained machine. 

Furthermore, we need a kind of ``Gramian matrix'' $\cal I$ based on the one-dimensional modi $\varphi_k$, too. The element ${\cal I}_{ij}$ is given by $\phi_i\odot\phi_j$. This matrix represents the linearization of the identity transformation $id(a)=a$. Comparing $\cal T$ with $\cal I$, e.g., by $\oplus$-adding these two matrices or by analyzing the kernel of the sum of them like in ``spectral analysis'', provides information about the complexity of the transformation $\Theta$ in $\cal B'$.  

After applying algebraic algorithms to compute the $\oplus$-sum or to  extract the spectral information, the result can be mapped back by $\rho$ to readable texts. 

\subsection{An example}

Let us analyze the transformation $\theta$, which takes away all metaphors from a given text. Then this transformation $\theta$ has a corresponding transformation in $\cal B'$. For simplicity, we use the four-digits representation of Sec.~\ref{sec:stone}. In this representation $\Theta$ just turns the first digit into $1$. For the computation of the matrices $\cal T$ and $\cal I$, we use again the four pexts of Sec.~\ref{sec:example}
$$X_1=1100, X_2=0111, X_3=1001, \text{ and } X_4=0011.$$
The projection of the identity mapping has already be computed in Sec.~\ref{sec:example} (denoted as $\cal M$):
$$
{\cal I}=
\begin{pmatrix}
1100 & 0100 & 1000 & 0000 \\
0100 & 0111 & 0001 & 0011 \\
1000 & 0001 & 1001 & 0001 \\
0000 & 0011 & 0001 & 0011 
\end{pmatrix}.
$$
After transformation we have:
$$\Theta(X_1)=1100, \Theta(X_2)=1111, \Theta(X_3)=1001, \text{ and } \Theta(X_4)=1011.$$
This can be used to do the Galerkin projection of the transformation $\Theta$:
$$
{\cal T}=
\begin{pmatrix}
1100 & 1100 & 1000 & 1000 \\
0100 & 0111 & 0001 & 0011 \\
1000 & 1001 & 1001 & 1001 \\
0000 & 0011 & 0001 & 0011 
\end{pmatrix}.
$$
Adding these two matrices leads to
$$
{\cal T}\oplus{\cal I}=
\begin{pmatrix}
0000 & 1000 & 0000 & 1000 \\
0000 & 0000 & 0000 & 0000 \\
0000 & 1000 & 0000 & 1000 \\
0000 & 0000 & 0000 & 0000 
\end{pmatrix}.
$$
All eigenmodi of the generalized eigenvalue problem correspond to the kernel of this matrix. The kernel of this matrix is very rich. This shows a low complexity of this transformation $\theta$ with regard to this special ``dialect'' $\cal B'$. A ``large'' kernel indicates a low complexity of the mapping $\Theta$. There is an invariant sub-module of this matrix spanned by the modi $(0000,1111,0000,0000)^T$, $(0000,0000,0000,1111)^T$, and $(1111,0000,1111,0000)^T$. However, all elements of this sub-module are nil-potent.

\section{Infinite Boolean Rings and Complexity}\label{sec:NP}

We have already seen in the last section, that the "injectivity" of the transformation $\Theta$ is a measure of its complexity. However, $\Theta$ is not an arbritary transformation. We want to study the {\em  computational} complexity of comparative text analysis. Algebraic methods for computing suitable elements of matrix kernels play an important role in these transformations. {If it is possible to formulate problems of comparative text analysis in terms of algebraic problems, then it might be interesting to know, whether these algebraic problems are "easy to solve" from a computational point of view.}

\subsection{Illustrative example with low complexity}
{Comparative text analysis is not only about the texts, it is also about the innovation and ideas of authors and their scribal skills, like ``Do these texts really include some original ideas?'' or like ``Is there a group of authors, which always only takes over the ideas from other authors?''.} We will now show why this might lead to a very complex question of comparative text analysis and that we are still lacking efficient algorithms to answer it.  

In order to illustrate, how this problem of comparative text analysis might look like, we come back to the title of this article. {If someone wants to show, that all common characteristics of modern crime novels are not new, this person could prove this hypothesis by the following argumentation: "Look, I have two stacks of crime novels. All crime novels on the left stack have the following common characteristics: ... and ... and .... Now look at my right stack of crime novels. You will see, that all these characteristics can be found here, too. This characteristic can be found, for example, in this novel and that characteristic in that novel. And so on."}

Let us assume, the crime novels (of course their sets of essays are meant) on the left stack are denoted as $X_1,\ldots , X_k$ and the right stack is $X_{k+1},\ldots, X_n$, where $1<k<n$, then working out {\em all} common characteristics of the left stack is given by the product $L=X_1 \odot \ldots \odot X_k$. Now it is claimed, that $L$ is "covered" by the "union" of the characteristics of the right stack. How to express this in terms of algebra? One of De Morgan's laws \cite{DeMorgan} is: the intersection of the complements of given sets is equal to the complement of the union of these sets. Thus, one has to compute the product $$R=(1\oplus X_{k+1})\odot \ldots \odot (1\oplus X_n).$$ This product provides {\em all} characteristics which are {\em not} represented in the right stack. Now, the intended proof about the missing innovation of crime novel writers is given by showing that $L\odot R=0$.

\subsection{Increasing complexity}
Although it might be complicated to work out all commonalities of modern crime novels $L$ or even the "non-characteristics" of old ones $R$ in order to prove the illustrated hypothesis, this is not a very complex problem from an algebraic point of view. Computing the product of given expressions does not have a high complexity. This kind of problem is classified as being of complexity ${\cal P}_{\cal B}$. If the number of input texts is $n$, I need to do at most $n$ multiplications $\odot$ and at most $(n-1)$ sums $\oplus$, thus, at most $(2n-1)$ algebraic operations in total for computing $L\odot R$. $2n-1$ is a polynomial of $n$. ${\cal P}_{\cal B}$ is the class of problems which can be solved with a deterministic algorithm (on Boolean rings $\cal B$) and polynomial cost depending on the number of input variables $n$. 
 
The illustrative example, however, leads to a highly complex problem. Imagine, someone gives to you a number $n$ of texts (need not be crime novels). Your task is now to decide which text goes to which stack, such that $L\odot R=0$, or to show that a stack-assignment like this is impossible. One way of solving this problem would be: check all possible $2^n$ assignments and for each assignment check (with less than $2n$ operations) whether it provides a feasible solution ($L\odot R=0$?) or not. However, this ineffective algorithm would need something like $2n\cdot2^n$ operations, which is not a polynomial in $n$ anymore. Under some circumstances, trial-and-error could be a good strategy. Thus, we try a non-deterministic algorithm. We simply randomly assign the texts to the two stacks and check whether $L\odot R=0$. For every guess we need $2n$ operations at most. This type of problems is classified as ${\cal NP}_{\cal B}$: A non-deterministic algorithm needs polynomial cost for checking whether the guessed solution is correct.  

\subsection{Searching for an algorithm}
The important question now is, whether there really does not exist {\em any} deterministic algorithm to solve such ${\cal NP}_{\cal B}$ problems with polynomial costs. And this question is so important, that it is one of the millenium problems of mathematics \cite{millenium}. You would get a reward of 1 million dollar, if you are the first to know the answer to the question for real numbers (not for Boolean rings): Is ${\cal P}\not={\cal NP}$?  

Mihai Prunescu claimed 2003 that he can show ${\cal P}_{\cal B}\not={\cal NP}_{\cal B}$ for infinite Boolean rings \cite{DBLP:journals/mlq/Prunescu03}. Although, this proof would be very important for mathematics, his article has only very few citations. At ISLA 2014, the 5th Indian School on Logic and its Application held in Tezpur University in Idia, J.A. Makowsky explicitly formulated the mathematical problem that Prunescu has used to get his complexity result \cite{ISLA2014}. Prunescu's problem is the "zero divisor problem in Boolean rings". It is the same as the illustrated  stack-assignment problem of comparative text analysis. Thus, we still lack an efficient algorithm to solve that problem, and there are good reasons to believe that an efficient deterministic  algorithm for ``creating the two stacks'' simply does not exist. {It is still a hard problem to judge about the originality of text writing.}

\section{Conclusion}

From a mathematical point of view, it is difficult to study the complexity of computer-based text analysis. Text analysis seems to be an ill-posed problem. In order to make it at least ``solvable'' for a machine, we shouldn't withhold information from the machine that we use ourselves to carry out text analyzes. But this is not enough. Still we have the problem of instability and non-uniqueness. In order to be able to resolve this situation in this manuscript, we basically allow everything as "possible text analysis" that anyone could write about the text at some point. There are no discussions, no arguments between different points of view. We are like collectors who simply accept everything without comment.  

Whereas mathematicians are stricter when it comes to comparative text analysis. Here we sort out. Not everything that we found remains in our collection when we compare it with the results of other text analyzes. There are clearly formulated algebraic correlations and mathematical insights in this area. We can formalize problems of which one can ask about their predictability and their computational complexity. Unfortunately, these mathematical problems described seem to have nothing to do with the work of a literary scholar who is asked to compare concrete texts with one another.

The need for a deep text analysis comes into play through the back door. If we try to break down the incredibly high-dimensional information about "yes, that works ... no, that does not work as a TA" into a low-dimensional vector, then the algebraic operations of comparative text analysis are suddenly complicated and have to be learned or trained. And this learning does not work without trying it out on very specific texts guided by a human teacher. When we teach our four machines, we expect their test runs to output meaningful products from a comparative text analysis. And each time the machine components are reassembled (we ask a new question), we expect (different) meaningful answers. With every test run, the machine learns what really matters to us when it comes to analyzing the texts.

Don't we humans learn in a similar way? We learn by refining our classification into categories. When we are asked to describe what a hare looks like, we probably start by mentioning its long ears. As a child, we may have called a hare a dog at some point and been told that this is an incorrect classification. Then we had to learn that not the four paws are good as criteria, but the long ears.

\paragraph{Acknowledgement.} This conceptual manuscript has partially been funded by the excellence center MATH+ via its project ``The Evolution of Ancient Egyptian -- Quantitative and Non-Quantitative Mathematical Linguistics".  

Ralph Birk, Tonio Sebastian Richter, and Marcus Weber (three of the PIs of the project) teamed up to work on clustering of ancient texts two years ago. Results of the many discussions and of concrete suggestions during this time entered this manuscript. Ralph Birk gave us a clearer understanding of literary studies and rephrased some of our arguments in this manuscript. We are very thankful. Together with Konstantin Fackeldey (the fourth PI) we mathematically reshaped the project such that it has been successfully granted within MATH+. Robert Julian Rabben and Tamaz Amiranashvili contributed to the discussion of possibly constructing a neural network doing the CTA - a kind of humanchine.

\bibliographystyle{plain}
\bibliography{references}

\appendix
\section{For illustration of $\odot$, $\oplus$, $\phi(f(\cdot))$, and $\rho$}\label{app:softw}

Put this file 'phi.m' into the corresponding working directory of Octave \textregistered. 

\begin{verbatim}
classdef phi
  properties
    value=0;
  endproperties
  
  methods
    function retval = phi(input1)
         retval.value=exp(bin2dec(input1)/1000)-1;
    endfunction
    function retval = mtimes(input1, input2)
         num1=round(log(input1.value+1)*1000);
         num2=round(log(input2.value+1)*1000);
         retval=phi('0');
         retval.value=exp(bitand(num1, num2)/1000)-1;
    endfunction
    function retval = plus(input1, input2)
         num1=round(log(input1.value+1)*1000);
         num2=round(log(input2.value+1)*1000);
         retval=phi('0');
         retval.value=exp(bitxor(num1, num2)/1000)-1;
    endfunction
    function rc = rho(input1)
       num1=round(log(input1.value+1)*1000);
       if (num1==0)
         retval=0;
       else
         retval=1;
         while(mod(num1,2)==0)
           retval=retval+1;
           num1=num1/2;
         end
       end
       switch (retval)
         case 0 rc='NULL';
         case 1 rc='In the end the "evil" is punished.';
         case 2 rc='The fairy tale ends with a wedding.';
         case 3 rc='A "wicked stepmother".';
         case 4 rc='The main character is a human.';
         case 5 rc='The main character is noble by birth.';
       end  
     endfunction
  endmethods
endclassdef
\end{verbatim}

After doing so: Execute this script 'boolscript.m' in the corresponding working directory of Octave \textregistered.
\begin{verbatim}
clear

% criteria 
% 5 Noble 
% 4 Human 
% 3 "wicked stepmother" 
% 2 happy end 
% 1 punishment 

% Sleeping Beauty 
m1=phi_('11010')
% Snow White 
m2=phi_('11111')
% Cinderella 
m3=phi_('01111')
% Hans in Luck 
m4=phi_('01000')
% The Wolf and the Seven Little Goats 
m5=phi_('00001')

% compute the unity
one=m1+m2+m1*m2;
one=one+m3+one*m3;
one=one+m4+one*m4;
one=one+m5+one*m5;

\end{verbatim}
\end{document}